\documentclass{article}

\usepackage[nonatbib,final]{nips_2016}

\usepackage[utf8]{inputenc} 
\usepackage[T1]{fontenc}    
\usepackage{hyperref}       
\usepackage{url}            
\usepackage{booktabs}       
\usepackage{amsfonts}       
\usepackage{nicefrac}       
\usepackage{microtype}      

\usepackage{graphicx}
\usepackage{amsmath,amssymb} 
\usepackage{color}

\usepackage{times}
\usepackage{url}
\usepackage{array}
\usepackage{multirow}
\usepackage{rotating}
\usepackage{arydshln}
\usepackage{bm}
\usepackage{balance}

\usepackage{xspace}
\newcommand*{\eg}{e.g.\@\xspace}
\newcommand*{\ie}{i.e.\@\xspace}

\makeatletter
\newcommand*{\etc}{%
    \@ifnextchar{.}%
        {etc}%
        {etc.\@\xspace}%
}
\makeatother

\renewcommand\footnotemark{}

\title{Deep Fully-Connected Networks for Video Compressive Sensing}

\author{
  Michael~Iliadis$^\ast$ \\  
  Northwestern University, EECS\\  
  \texttt{miliad@u.northwestern.edu} \\
  \And
   Leonidas Spinoulas$^\ast$ \\
   Northwestern University, EECS \\  
   \texttt{leonisp@u.northwestern.edu} \\
   \AND
   Aggelos K. Katsaggelos \\
   Northwestern University, EECS \\  
   \texttt{aggk@eecs.northwestern.edu } \\
}
\thanks{$^*$Indicates equal contribution.}

\begin{document}

\maketitle

\begin{abstract}
In this work we present a deep learning framework for video compressive sensing. The proposed formulation enables recovery of video frames in a few seconds at significantly improved reconstruction quality compared to previous approaches. Our investigation starts by learning a linear mapping between video sequences and corresponding measured frames which turns out to provide promising results. We then extend the linear formulation to deep fully-connected networks and explore the performance gains using deeper architectures. Our analysis is always driven by the applicability of the proposed framework on existing compressive video architectures. Extensive simulations on several video sequences document the superiority of our approach both quantitatively and qualitatively. Finally, our analysis offers insights into understanding how dataset sizes and number of layers affect reconstruction performance while raising a few points for future investigation.

Code is available at Github: \url{https://github.com/miliadis/DeepVideoCS}  
\end{abstract}

\section{Introduction}

The subdivision of time by motion picture cameras, the frame rate, limits the temporal resolution of a camera system. Even though frame rate increase above $30$~Hz may be imperceptible to human eyes, high speed motion picture capture has long been a goal in scientific imaging and cinematography communities. Despite the increasing availability of high speed cameras through the reduction of hardware prices, fundamental restrictions still limit the maximum achievable frame rates.

Video compressive sensing (CS) aims at increasing the temporal resolution of a sensor by incorporating additional hardware components to the camera architecture and employing powerful computational techniques for high speed video reconstruction. The additional components operate at higher frame rates than the camera's native temporal resolution giving rise to low frame rate multiplexed measurements which can later be decoded to extract the unknown observed high speed video sequence. Despite its use for high speed motion capture~\cite{Llull2015}, video CS also has applications to coherent imaging (\eg, holography) for tracking high-speed events~\cite{Wang2017} (\eg, particle tracking, observing moving biological samples). The benefits of video CS are even more pronounced for non-visible light applications where high speed cameras are rarely available or prohibitively expensive (\eg, millimeter-wave imaging, infrared imaging)~\cite{Babacan2011,Chen2015}.

Video CS comes in two incarnations, namely, spatial CS and temporal CS. Spatial video CS architectures stem from the well-known single-pixel-camera~\cite{Duarte2008}, which performs spatial multiplexing per measurement, and enable video recovery by expediting the capturing process. They either employ fast readout circuitry to capture information at video rates~\cite{Chen2014} or parallelize the single-pixel architecture using multiple sensors, each one responsible for sampling a separate spatial area of the scene~\cite{Chen2015,Wang2015}.

\begin{figure}[!t]
\centering
\includegraphics{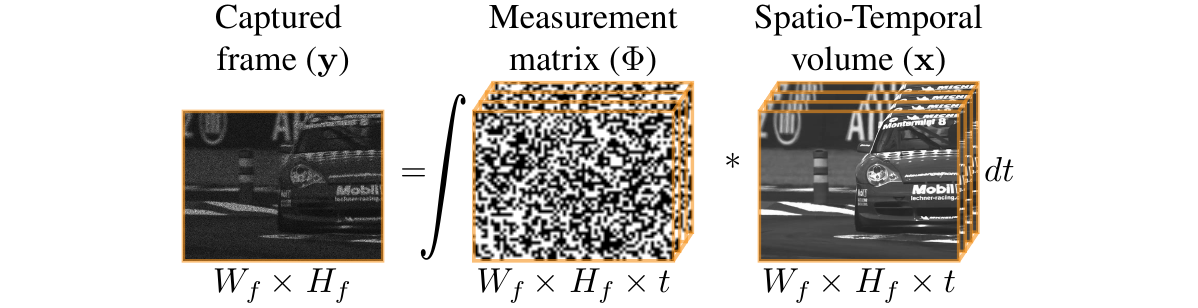}
\caption{Temporal compressive sensing measurement model.}
\label{fig:measurementModel}
\end{figure}

In this work, we focus on temporal CS where multiplexing occurs across the time dimension. Figure~\ref{fig:measurementModel} depicts this process, where a spatio-temporal volume of size $W_{f}\times H_{f} \times t = N_{f}$ is modulated by $t$ binary random masks during the exposure time of a single capture, giving rise to a coded frame of size $W_{f} \times H_{f} = M_{f}$.

We denote the vectorized versions of the unknown signal and the captured frame as ${\bf x}: N_{f}\times 1$ and ${\bf y}: M_{f} \times 1$, respectively. Each vectorized sampling mask is expressed as ${\bm \phi}_{1}, \dots, {\bm \phi}_{t}$ giving rise to the measurement model
\begin{equation}
{\bf y} = \Phi {\bf x},
\label{eq:measurementModel}
\end{equation}
where $\Phi = \left[ diag({\bm \phi}_{1}), \dots, diag({\bm \phi}_{t})\right] : M_{f} \times N_{f}$ and $diag(\cdot)$ creates a diagonal matrix from its vector argument.

Various successful temporal CS architectures have been proposed. Their differences mainly involve the implementation of the random masks on the optical path (\ie, the measurement matrix in Figure~\ref{fig:measurementModel}). Digital micromirror devices (DMD), spatial light modulators (SLM) and liquid crystal on silicon (LCoS) were used in~\cite{Chen2015,Wang2015,Gao2014,Liu2013,Reddy2011} while translating printed masks were employed in~\cite{Koller2015,Llull2013b}. Moreover, a few architectures have eliminated additional optical elements by directly programming the chip's readout mode through hardware circuitry modifications~\cite{Fernandez-Cull2014,Orchard2012,Spinoulas2015}.

Despite their reasonable performance, temporal CS architectures lack practicality. The main drawback is that existing reconstruction algorithms (\eg, using sparsity models~\cite{Chen2015,Holloway2012}, combining sparsity and dictionary learning~\cite{Liu2013} or using Gaussian mixture models~\cite{Yang2015,Yang2014}) are often too computationally intensive, rendering the reconstruction process painfully slow. Even with parallel processing, recovery times make video CS prohibitive for modern commercial camera architectures.

In this work, we address this problem by employing deep learning and show that video frames can be recovered in a few seconds at significantly improved reconstruction quality compared to existing approaches.

Our contributions are summarized as follows: 
\begin{enumerate}
\item We present the first deep learning architecture for temporal video CS reconstruction approach, based on fully-connected neural networks, which learns to map directly temporal CS measurements to video frames. For such task to be practical, a measurement mask with a repeated pattern is proposed.
\item We show that a simple linear regression-based approach learns to reconstruct video frames adequately at a minimal computational cost. Such reconstruction could be used as an initial point to other video CS algorithms.
\item The learning parading is extended to deeper architectures exhibiting reconstruction quality and computational cost improvements compared to previous methods.
\end{enumerate}

\section{Motivation and Related Work}

Deep learning~\cite{LeCun2015} is a burgeoning research field which has demonstrated state-of-the-art performance in a multitude of machine learning and computer vision tasks, such as image recognition~\cite{He2015} or object detection~\cite{Pinheiro2015}.

In simple words, deep learning tries to mimic the human brain by training large multi-layer neural networks with vast amounts of training samples, describing a given task. Such networks have proven very successful in problems where analytical modeling is not easy or straightforward (\eg, a variety of computer vision tasks~\cite{Krizhevsky2012,Lecun1998}).

The popularity of neural networks in recent years has led researchers to explore the capabilities of deep architectures even in problems where analytical models often exist and are well understood (\eg, restoration problems~\cite{Burger2012,Schuler2013,Xie2012}). Even though performance improvement is not as pronounced as in classification problems, many proposed architectures have achieved state-of-the-art performance in problems such as deconvolution, denoising, inpainting, and super-resolution.

More specifically, investigators have employed a variety of architectures: deep fully-connected networks or multi-layer perceptrons (MLPs)~\cite{Burger2012,Schuler2013}; stacked denoising auto-encoders (SDAEs)~\cite{Xie2012,Agostinelli2013,Fleet2014,Vincent2010}, which are MLPs whose layers are pre-trained to provide improved weight initialization; convolutional neural networks (CNNs) \cite{Wang2015,Sun2015,Dong2015,Lecun1989,Ren2015,Li2014} and recurrent neural networks (RNNs)~\cite{Yan2015}.

Based on such success in restoration problems, we wanted to explore the capabilities of deep learning for the video CS problem. However, the majority of existing architectures involve outputs whose dimensionality is smaller than the input (\eg, classification) or have the same size (\eg, denoising/deblurring). Hence, devising an architecture that estimates $N_{f}$ unknowns, given $M_{f}$ inputs, where $M_{f} \ll N_{f}$ is not necessarily straightforward.

Two recent studies, utilizing SDAEs~\cite{Mousavi2015} or CNNs~\cite{Kulkarni2016}, have been presented on spatial CS for still images exhibiting promising performance. Our work constitutes the first attempt to apply deep learning on temporal video CS. Our approach differs from prior 2D image restoration architectures~\cite{Burger2012,Schuler2013} since we are recovering a 3D volume from 2D measurements.

\section{Deep Networks for Compressed Video}
\subsection{Linear mapping}

We started our investigation by posing the question: can training data be used to find a linear mapping $W$ such that ${\bf x} = W{\bf y}$? Essentially, this question asks for the inverse of $\Phi$ in equation~\eqref{eq:measurementModel} which, of course, does not exist. Clearly, such a matrix would be huge to store but, instead, one can apply the same logic on video blocks~\cite{Liu2013}.

We collect a set of training video blocks denoted by ${\bf x}_i$, $i \in \mathbb{N}$ of size $w_p \times h_p \times t = N_{p}$. Therefore, the measurement model per block is now ${\bf y}_{i} = \Phi_{p}{\bf x}_{i}$ with size $M_p \times 1$, where $M_p = w_{p}\times h_{p}$ and $\Phi_{p}$ refers to the corresponding measurement matrix per block.

Collecting a set of $N$ video blocks, we obtain the matrix equation
\begin{equation}
Y = \Phi_{p} X,
\label{eq:matrixMapping}
\end{equation}
where $Y = \left[ {\bf y}_{1},\dots,{\bf y}_{N}\right]$, $X = \left[{\bf x}_{1},\dots,{\bf x}_{N}\right]$ and $\Phi_{p}$ is the same for all blocks. The linear mapping $X = W_{p} Y$ we are after can be calculated as
\begin{equation}
\min_{W_{p}} \left\lVert X - W_{p}Y\right\rVert_{2}^{2} \rightarrow W_{p} = \left( XY^T\right)\left(YY^T\right)^{-1},
\label{eq:mappingSolution}
\end{equation}
where $W_{p}$ is of size $N_{p} \times M_{p}$.

\begin{figure}[!t]
\centering
\includegraphics{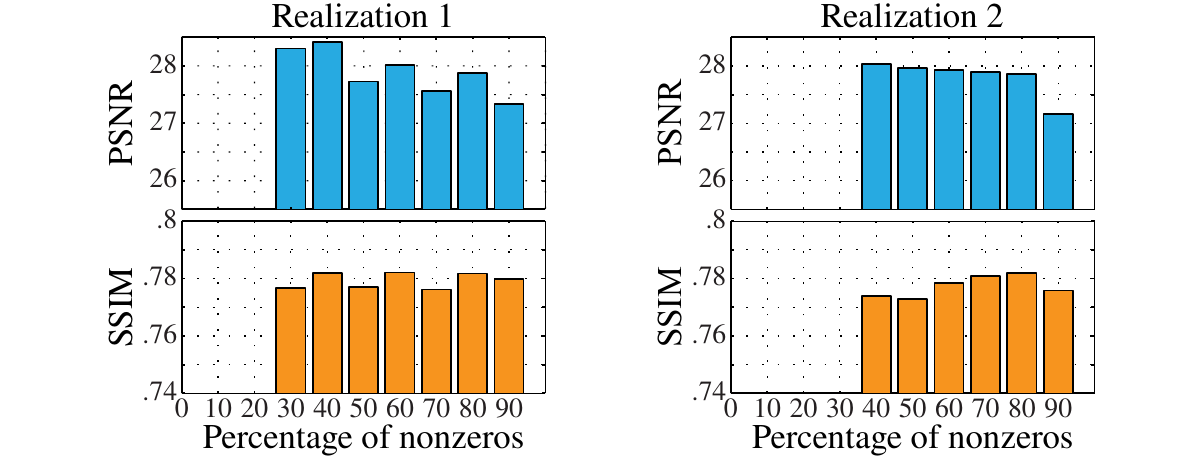}
\caption{Average reconstruction performance of linear mapping for $14$ videos (unrelated to the training data), using measurement matrices $\Phi_p$ with varying percentages of nonzero elements.}
\label{fig:Wcomparison}
\end{figure}

Intuitively, such an approach would not necessarily be expected to even provide a solution due to ill-posedness. However, it turns out that, if $N$ is sufficiently large and the matrix $\Phi_p$ has at least one nonzero in each row (\ie, sampling each spatial location at least once over time), the estimation of ${\bf x}_i$'s by the ${\bf y}_i$'s provides surprisingly good performance. 

Specifically, we obtain measurements from a test video sequence applying the same $\Phi_{p}$ per video block and then reconstruct all blocks using the learnt $W_{p}$. Figure~\ref{fig:Wcomparison} depicts the average peak signal-to-noise ratio (PSNR) and structural similarity metric (SSIM)~\cite{Wang2004} for the reconstruction of $14$ video sequences using $2$ different realizations of the random binary matrix $\Phi_p$ for varying percentages of nonzero elements. The empty bars for $10-20\%$ and $10-30\%$ of nonzeros in realizations $1$ and $2$, respectively, refer to cases when there was no solution due to the lack of nonzeros at some spatial location. In these experiments $w_{p} \times h_{p} \times t$ was selected as $8 \times 8 \times 16$ simulating the reconstruction of $16$ frames by a single captured frame and $N = 10^{6}$.

\subsection{Measurement Matrix Construction}
\label{subsec:mask_construction}
\begin{figure}[!t]
\centering
\includegraphics{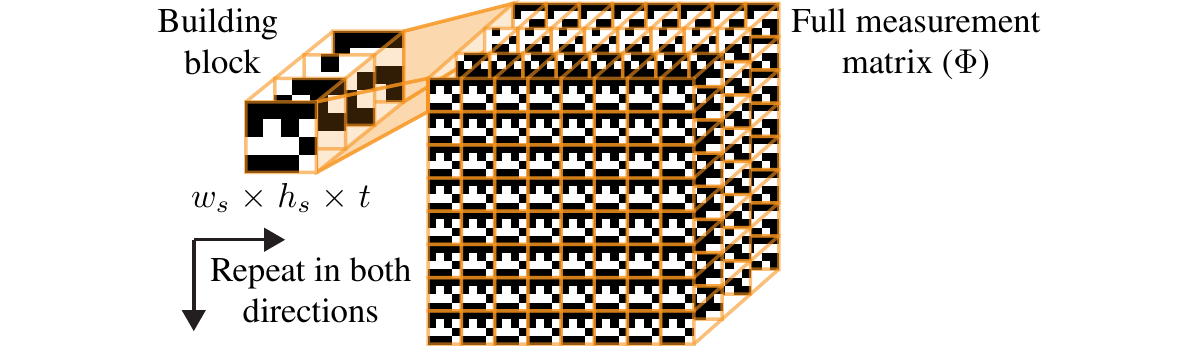}
\caption{Construction of the proposed full measurement matrix by repeating a three dimensional random array (building block) in the horizontal and vertical directions.}
\label{fig:maskConstruction}
\end{figure}

Based on the performance in Figure~\ref{fig:Wcomparison}, investigating the extension of the linear mapping in $\eqref{eq:mappingSolution}$ to a nonlinear mapping using deep networks seemed increasingly promising. In order for such an approach to be practical, though, reconstruction has to be performed on blocks and each block must be sampled with the same measurement matrix $\Phi_p$. Furthermore, such a measurement matrix should be realizable in hardware. Hence we propose constructing a $\Phi$ which consists of repeated identical building blocks of size $w_{s} \times h_{s} \times t$, as presented in Figure~\ref{fig:maskConstruction}. Such a matrix can be straightforwardly implemented on existing systems employing DMDs, SLMs or LCoS~\cite{Chen2015,Wang2015,Gao2014,Liu2013,Reddy2011}. At the same time, in systems utilizing translating masks~\cite{Koller2015,Llull2013b}, a repeated mask can be printed and shifted appropriately to produce the same effect.

In the remainder of this paper, we select a building block of size $w_{s} \times h_{s} \times t = 4 \times 4 \times 16$ as a random binary matrix containing $50\%$ of nonzero elements and set $w_{p} \times h_{p} \times t = 8 \times 8 \times 16$, such that $N_{p} = 1024$ and $M_{p} = 64$. Therefore, the compression ratio is $1/16$. In addition, for the proposed matrix $\Phi$, each $4 \times 4 \times 16$ block is the same allowing reconstruction for overlapping blocks of size $8 \times 8 \times 16$ with spatial overlap of $4 \times 4$. Such overlap can usually aid at improving reconstruction quality. The selection of $50\%$ of nonzeros was just a random choice since the results of Figure~\ref{fig:Wcomparison} did not suggest that a specific percentage is particularly beneficial in terms of reconstruction quality.

\subsection{Multi-layer Network Architecture}
\begin{figure*}[!t]
\begin{center}
\includegraphics{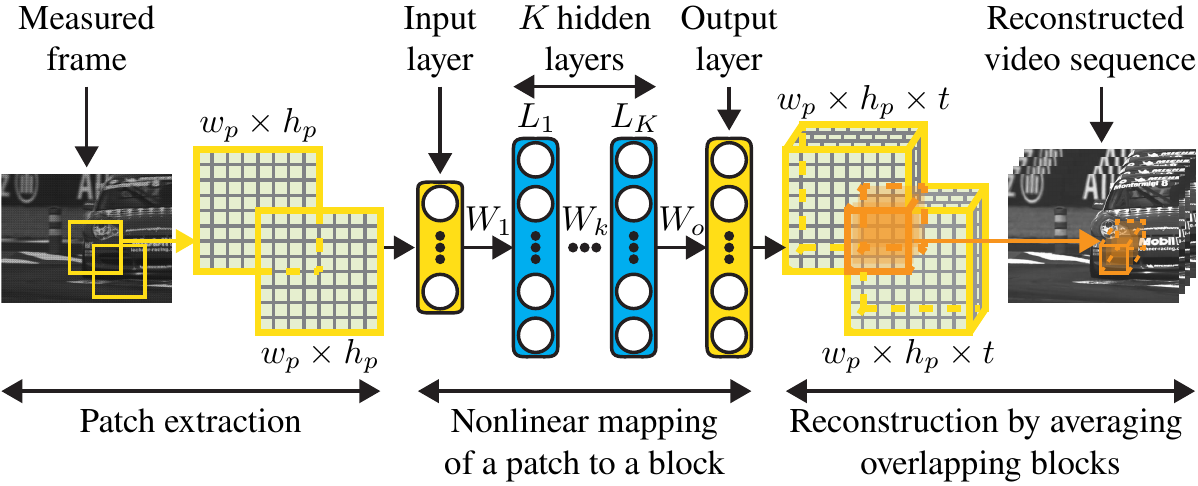}
\end{center}
\caption{Illustration of the proposed deep learning architecture for video compressive sensing.}
\label{fig:network}
\end{figure*}

In this section, we extend the linear formulation to MLPs and investigate the performance in deeper structures. \newline \newline {\bf Choice of Network Architecture.} We consider an end-to-end MLP architecture to learn a nonlinear function $f(\cdot)$ that maps a measured frame patch ${\bf y}_{i}$ via several hidden layers to a video block ${\bf x}_{i}$, as illustrated in Figure~\ref{fig:network}. The MLP architecture was chosen for the problem of video CS reconstruction due to the following two considerations;

\begin{enumerate}  
\item The first hidden layer should be a fully-connected layer that would provide a 3D signal from the compressed 2D measurements. This is necessary for temporal video CS as in contrast to the super-resolution problem (or other related image reconstruction problems) where a low-resolution image is given as input, here we are given CS encoded measurements. Thus, convolution does not hold and therefore a convolutional layer cannot be employed as a first layer.  
\item Following that, one could argue that the subsequent layers could be 3D Convolutional layers~\cite{Tran2015}. Although that would sound reasonable for our problem, in practice, the small size of blocks used in this paper ($8 \times 8 \times 16$) do not allow for convolutions to be effective. Increasing the size of blocks to $32 \times 32 \times 16$, so that convolutions can be applied, would dramatically increase the network complexity in 3D volumes such as in videos. For example, if we use a block size of $32\times32$ as input, the first fully-connected layer would contain $(32 \times 32 \times 16) \times (32 \times 32)= 16,777,216$ parameters! Besides, such small block sizes ($8 \times 8 \times 16$) have provided good reconstruction quality in dictionary learning approaches used for CS video reconstruction~\cite{Liu2013}. It was shown that choosing larger block sizes led to worse reconstruction quality.
\end{enumerate}
Thus, MLPs (\ie, apply fully-connected layers for the entire network) were considered more reasonable in our work and we found that when applied to $8 \times 8 \times 16$ blocks they capture the motion and spatial details of videos adequately.

It is interesting to note here that another approach would be to try learning the mapping between ${\hat {\bf x}_i} = \Phi_p^T {\bf y}_i$ and ${\bf x}_i$, since matrix $\Phi_p$ is known~\cite{Mehta17}. Such approach could provide better pixel localization since $\Phi_p^T{\bf y}$ places the values in ${\bf y}$ in the corresponding pixel locations that were sampled to provide the summation in the $t$ direction. However, such an architecture would require additional weights between the input and the first hidden layer since the input would now be of size ($8\times 8 \times 16$) instead of ($8 \times 8$). Such approach was tested and resulted in almost identical performance, albeit with a higher computational cost, hence it is not presented here.
\newline \newline {\bf Network Architecture Design.}
As illustrated in Figure~\ref{fig:network}, each hidden layer $L_k$, $k = 1, \dots, K$ is defined as
\begin{equation}
h_k({\bf y}) = \sigma({\bf b}_{k} +  W_{k} {\bf y}), 
\end{equation}
where ${\bf b}_k \in \mathbb{R}^{N_p}$ is the bias vector and $W_{k}$ is the output weight matrix, containing linear filters. $W_1 \in \mathbb{R}^{N_p \times M_p}$ connects ${\bf y}_i$ to the first hidden layer, while for the remaining hidden layers, ${W}_{2-K}\in \mathbb{R}^{N_p \times N_p}$. The last hidden layer is connected to the output layer via ${\bf b}_{o} \in \mathbb{R}^{N_p}$ and $W_o \in \mathbb{R}^{N_p \times N_p} $ without nonlinearity. The non-linear function $\sigma(\cdot)$ is the rectified linear unit (ReLU)~\cite{Nair2010} defined as, $\sigma(y) = \max(0,y)$. In our work we considered two different network architectures, one with $K=4$ and another with $K=7$ hidden layers.

To train the proposed MLP, we learn all the weights and biases of the model. The set of parameters is denoted as $\theta = \left\{{\bf b}_{1-K},{\bf b}_o,W_{1-K},W_o \right\}$ and is updated by the backpropagation algorithm~\cite{Rumelhart1988} minimizing the quadratic error between the set of training mapped measurements $f({\bf y}_i;\theta)$ and the corresponding video blocks ${\bf x}_i$. The loss function is the mean squared error (MSE) which is given by 
\begin{equation}
L(\theta) = \frac{1}{N}\sum_{i = 1}^N{\left\Vert f({\bf y}_i;\theta) - {\bf x}_i\right\Vert^2_2}. 
\end{equation}
The MSE was used in this work since our goal is to optimize the PSNR which is directly related to the MSE.


\begin{figure*}[!t]
\centering
\includegraphics{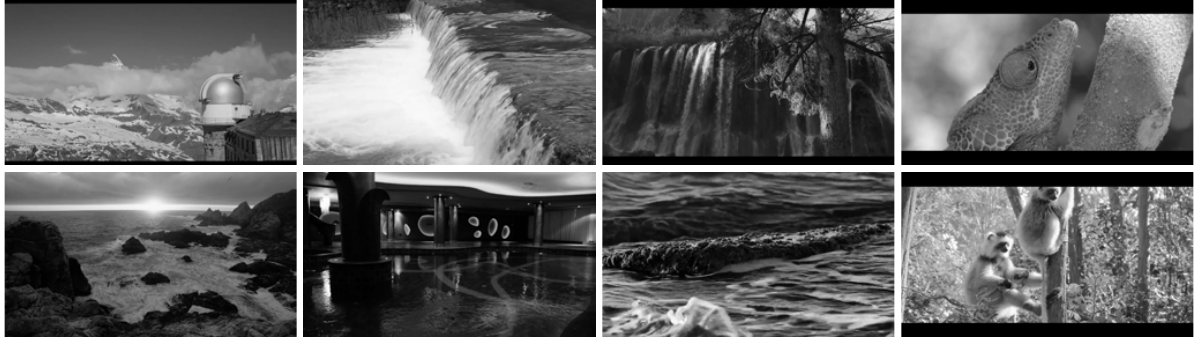}
\caption{Example frames from the video sequences used for training.}
\label{fig:trainingFrames}
\end{figure*}

\section{Experiments}

We compare our proposed deep architecture with state-of-the-art approaches both quantitatively and qualitatively. The proposed approaches are evaluated assuming noiseless measurements or under the presence of measurement noise. Finally, we investigate the performance of our methods under different network parameters (\eg, number of layers) and size of training samples. The metrics used for evaluation were the PSNR and SSIM.

\subsection{Training Data Collection}
\label{subsec:training_data}

For deep neural networks, increasing the number of training samples is usually synonymous to improved performance. We collected a diverse set of training samples using $400$ high-definition videos from Youtube, depicting natural scenes. The video sequences contain more than $10^5$ frames which were converted to grayscale. All videos are unrelated to the test set. We randomly extracted $10$ million video blocks of size $w_p \times h_p \times t$ while keeping the amount of blocks extracted per video proportional to its duration. This data was used as output while the corresponding input was obtained by multiplying each sample with the measurement matrix $\Phi_p$ (see subsection~\ref{subsec:mask_construction} for details). Example frames from the video sequences used for training are shown in Figure~\ref{fig:trainingFrames}.

\subsection{Implementation Details}

Our networks were trained for up to $4 \times 10^6$ iterations using a mini-batch size of $200$. We normalized the input per-feature to zero mean and standard deviation one. The weights of each layer were initialized to random values uniformly distributed in $(-1/\sqrt{s},1/\sqrt{s})$, where $s$ is the size of the previous layer~\cite{Xavier2010}. We used Stochastic Gradient Descent (SGD) with a starting learning rate of $0.01$, which was divided by $10$ after $3 \times 10^6$ iterations. The momentum was set to 0.9 and we further used $\ell_2$ norm gradient clipping to keep the gradients in a certain range. Gradient clipping is a widely used technique in recurrent neural networks to avoid exploding gradients~\cite{Pascanu2013}. The threshold of gradient clipping was set to $10$.

\subsection{Comparison with Previous Methods}
\label{subsec:comparison}

We compare our method with the state-of-the-art video compressive sensing methods:
\begin{itemize}
  \item GMM-TP, a Gaussian mixture model (GMM)-based algorithm~\cite{Yang2014}.
  \item MMLE-GMM, a maximum marginal likelihood estimator (MMLE), that maximizes the likelihood of the GMM of the underlying signals given only their linear compressive measurements~\cite{Yang2015}. 
\end{itemize}
For temporal CS reconstruction, data driven models usually perform better than standard sparsity-based schemes~\cite{Yang2015,Yang2014}. Indeed, both GMM-TP and MMLE-GMM have demonstrated superior performance compared to existing approaches in the literature such as Total-Variation (TV) or dictionary learning~\cite{Liu2013,Yang2015,Yang2014}, hence we did not include experiments with the latter methods.

In GMM-TP~\cite{Yang2014} we followed the settings proposed by the authors and used our training data (randomly selecting $20,000$ samples) to train the underlying GMM parameters. We found that our training data provided better performance compared to the data used by the authors. In our experiments we denote this method by GMM-$4$ to denote reconstruction of overlapping blocks with spatial overlap of $4 \times 4$ pixels, as discussed in subsection~\ref{subsec:mask_construction}. 

MMLE~\cite{Yang2015} is a self-training method but it is sensitive to initialization. A satisfactory performance is obtained only when MMLE is combined with a good starting point. In~\cite{Yang2015}, the GMM-TP~\cite{Yang2014} with full overlapping patches (denoted in our experiments as GMM-1) was used to initialize the MMLE. We denote the combined method as GMM-1+MMLE. For fairness, we also conducted experiments in the case where our method is used as a starting point for the MMLE.  

\begin{table*}[!t]
\caption{Average performance for the reconstruction of the first $32$ frames for $14$ video sequences using several methods. Maximum values are highlighted for each side (left/right) of the table. The time (at the bottom row) refers to the average time for reconstructing a sequence of $16$ frames using a single captured frame.}
\label{tb:psnr_table}
\begin{center}
\begin{small}
\resizebox{!}{5.3cm}{
\begin{tabular}{c|c||c|c|c|c||c|c|}
\cline{3-8}
 \multicolumn{2}{c|}{} &\multicolumn{6}{c|}{Reconstruction Method} \\ \hline

\multirow{2}{*}{\parbox{1.5cm}{\centering Video Sequence}} &  
\multirow{2}{*}{Metric}      & 
\multirow{2}{*}{W-10M}    &  
\multirow{2}{*}{FC7-10M} &
\multirow{2}{*}{GMM-4~\cite{Yang2014}}    &  
\multirow{2}{*}{GMM-1~\cite{Yang2015}}    & 
\multirow{2}{*}{\parbox{1.25cm}{\centering FC7-10M +MMLE}} &
\multirow{2}{*}{\parbox{1.8cm}{\centering GMM-1 +MMLE~\cite{Yang2015}}} \\
& &  & & & & & \\ \hline \hline

\multirow{2}{*}{\parbox{1cm}{\centering Electric Ball}}
& {\footnotesize PSNR}& $40.97$ & ${\bf 43.62}$ & $40.16$ & $40.27$ & \color{red}{${\bf 43.81}$} & $41.18$  \\  
\cdashline{2-8}
& {\footnotesize SSIM}& $0.9796$ & ${\bf 0.9882}$ & $0.9747$ & $0.9754$ & \color{red}{${\bf 0.9885}$} & $0.9802$    
\\ \hline

\multirow{2}{*}{Horse}   
& {\footnotesize PSNR} & $29.08$ & ${\bf 31.65}$ & $29.00$ & $29.20$ & \color{red}{${\bf 31.58}$} & $29.97$ \\       \cdashline{2-8}
& {\footnotesize SSIM}& $0.7869$ & ${\bf 0.8586}$ & $0.7747$ & $0.7803$ & \color{red}{${\bf 0.8556}$} & $0.8016$    
\\ \hline

\multirow{2}{*}{\parbox{1.0cm}{\centering Bow \& Arrow}}
&{\footnotesize PSNR}& $35.59$ & ${\bf 41.88}$ & $39.27$ & $40.06$ & \color{red}{${\bf 42.96}$} & $41.77$ \\      
\cdashline{2-8}
& {\footnotesize SSIM} &$0.9773$ & ${\bf 0.9885}$ & $0.9810$ & $0.9838$ & \color{red}{${\bf 0.9902}$} & $0.9881$    
\\ \hline

\multirow{2}{*}{Bus}  
& {\footnotesize PSNR} &$18.92$ & ${\bf 20.10}$ & $19.01$ & $19.20$ & \color{red}{${\bf 20.22}$} & $19.35$ \\     
\cdashline{2-8}
& {\footnotesize SSIM} & $0.4583$ & ${\bf 0.5316}$ & $0.4640$ & $0.4817$ & \color{red}{${\bf 0.5375}$} & $0.4815$    
\\ \hline

\multirow{2}{*}{Dogs}   
& {\footnotesize PSNR}& $35.64$ & ${\bf 42.40}$ & $38.03$ & $39.29$ & \color{red}{${\bf 43.50}$} & $42.39$ \\      
\cdashline{2-8}
& {\footnotesize SSIM} &$0.9712$ & ${\bf 0.9889}$ & $0.9739$ & $0.9796$ & \color{red}{${\bf 0.9929}$} & $0.9919$    
\\ \hline

\multirow{2}{*}{City}   
&{\footnotesize PSNR}& $22.28$ & ${\bf 23.29}$ & $22.39$ & $22.55$ & \color{red}{${\bf 23.16}$} & $22.55$ \\      
\cdashline{2-8}
& {\footnotesize SSIM} & $0.5127$ & ${\bf 0.6279}$ & $0.5196$ & $0.5302$ & \color{red}{${\bf 0.6408}$} & $0.5579$    
\\ \hline

\multirow{2}{*}{Crew}   
& {\footnotesize PSNR} & $29.74$ & ${\bf 32.48}$ & $29.68$ & $29.89$ & \color{red}{${\bf 33.35}$} & $30.42$ \\     
\cdashline{2-8}
& {\footnotesize SSIM} & $0.8362$ & ${\bf 0.8771}$ & $0.8371$ & $0.8450$ & \color{red}{${\bf 0.8943}$} & $0.8621$    
\\ \hline

\multirow{2}{*}{Filament}   
&{\footnotesize PSNR} &  $42.02$ & ${\bf 51.43}$ & $47.95$ & $49.33$ & \color{red}{${\bf 55.03}$} & $52.75$ \\    
\cdashline{2-8}
&{\footnotesize SSIM} & $0.9945$ & ${\bf 0.9974}$ & $0.9963$ & $0.9965$ & \color{red}{${\bf 0.9989}$} & $0.9988$   
\\ \hline

\multirow{2}{*}{Hammer}   
& {\footnotesize PSNR} &$31.11$ & ${\bf 38.04}$  & $34.45$ & $34.99$ & \color{red}{${\bf 38.59}$} & $36.68$ \\  
\cdashline{2-8}
&{\footnotesize SSIM} &$0.9304$ & ${\bf 0.9666}$  & $0.9360$ & $0.9423$ & \color{red}{${\bf 0.9696}$} & $0.9553$    
\\ \hline

\multirow{2}{*}{Football} 
& {\footnotesize PSNR} & $19.84$ & ${\bf 21.58}$ & $20.25$ & $20.46$ & \color{red}{${\bf 21.85}$} & $20.80$ \\      
\cdashline{2-8}
& {\footnotesize SSIM} &$0.4793$ & ${\bf 0.5642}$ & $0.5009$ & $0.5277$ & \color{red}{${\bf 0.5834}$} & $0.5378$    
\\ \hline

\multirow{2}{*}{Kayak} 
& {\footnotesize PSNR} & $26.49$ & ${\bf 30.46}$ & $27.41$ & $27.66$ & \color{red}{${\bf 30.55}$} & $28.74$ \\     
\cdashline{2-8}
&{\footnotesize SSIM}  & $0.7188$ & ${\bf 0.8128}$ & $0.7326$ & $0.7458$ & \color{red}{${\bf 0.8142}$} & $0.7638$    
\\ \hline

\multirow{2}{*}{Porsche} 
& {\footnotesize PSNR} & $26.17$ & ${\bf 29.52}$ & $26.14$ & $26.37$ & \color{red}{${\bf 30.15}$} & $27.45$ \\     
\cdashline{2-8}
& {\footnotesize SSIM}  & $0.9310$ & ${\bf 0.9640}$ & $0.9270$ & $0.9328$ & \color{red}{${\bf 0.9675}$} & $0.9491$    
\\ \hline

\multirow{2}{*}{Golf} 
& {\footnotesize PSNR} & $26.77$ & ${\bf 29.41}$ & $28.30$ & $28.58$ & \color{red}{${\bf 29.89}$} & $29.14$  \\    
\cdashline{2-8}
& {\footnotesize SSIM} & $0.9050$ & ${\bf 0.9401}$ & $0.9235$ & $0.9319$ & \color{red}{${\bf 0.9507}$} & $0.9440$    
\\ \hline

\multirow{2}{*}{Basketball} 
& {\footnotesize PSNR} & $22.35$ & ${\bf 25.15}$ & $22.80$ & $23.00$ & \color{red}{${\bf 25.53}$} & $23.66$ \\     
\cdashline{2-8}
& {\footnotesize SSIM} & $0.6412$ & ${\bf 0.7687}$ & $0.6640$ & $0.6868$ & \color{red}{${\bf 0.7860}$} & $0.7156$    
\\ \hline

& {\footnotesize Time}& \color{blue} $\sim1s$ & \color{blue} $\sim10s$ & \color{blue} $\sim100s$ & \color{blue} $\sim10^3s$ & \color{blue} $\sim10^3-10^4s$ & \color{blue} $\sim10^3-10^4s$ \\
\end{tabular}
}
\end{small}
\end{center}
\end{table*}

In our methods, a collection of overlapping patches of size $w_p \times h_p$ is extracted by each coded measurement of size $W_f \times H_f$ and subsequently reconstructed into video blocks of size $w_p \times h_p \times t$. Overlapping areas of the recovered video blocks are then averaged to obtain the final video reconstruction results, as depicted in Figure~\ref{fig:network}. The step of the overlapping patches was set to $4 \times 4$ due to the special construction of the utilized measurement matrix, as discussed in subsection~\ref{subsec:mask_construction}. 

We consider six different architectures:
\begin{itemize}
  \item W-10M, a simple linear mapping (equation~\eqref{eq:mappingSolution}) trained on $10\times10^6$ samples.
  \item FC4-1M, a $K=4$ MLP trained on $1 \times 10^6$ samples (randomly selected from our $10\times10^6$ samples).
  \item FC4-10M, a $K=4$ MLP trained on $10 \times 10^6$ samples.
   \item FC7-1M, a $K=7$ MLP trained on $1\times10^6$ samples (randomly selected from our $10\times10^6$ samples).
  \item FC7-10M, a $K=7$ MLP trained on $10\times10^6$ samples.  
  \item FC7-10M+MMLE, a $K=7$ MLP trained on $10 \times 10^6$ samples which is used as an initialization to the MMLE~\cite{Yang2015} method. 
\end{itemize}
Note that the subset of randomly selected $1$ million samples used for training FC4-1M and FC7-1M was the same.

Our test set consists of $14$ video sequences. They involve a set of videos that were used for dictionary training in~\cite{Liu2013}, provided by the authors, as well as the ``Basketball'' video sequence used by~\cite{Yang2015}. All video sequences are unrelated to the training set (see subsection~\ref{subsec:training_data} for details). For fair comparisons, the same measurement mask was used in all methods, according to subsection~\ref{subsec:mask_construction}. All code implementations are publicly available provided by the authors.

\subsection{Reconstruction Results}
\begin{figure}[!t]
\centering
\includegraphics{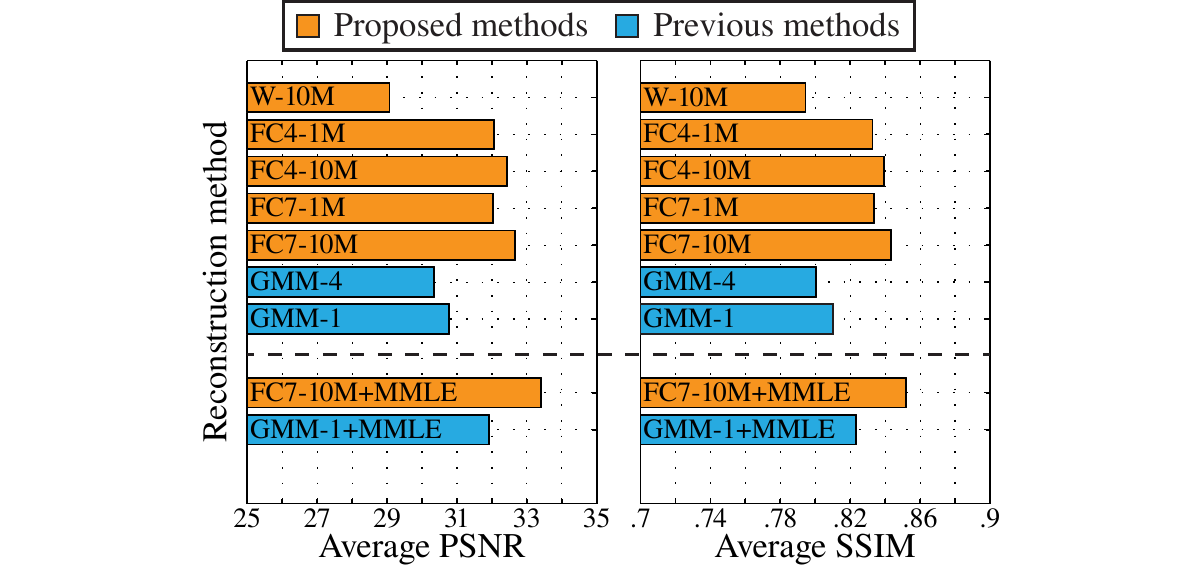}
\caption{Average PSNR and SSIM over all video sequences for several methods.}
\label{fig:allAlgorithmComparison}
\end{figure}

Quantitative reconstruction results for all video sequences with all tested algorithms are illustrated in Table~\ref{tb:psnr_table} and average performance is summarized in Figure~\ref{fig:allAlgorithmComparison}. The presented metrics refer to average performance for the reconstruction of the first $32$ frames of each video sequence, using $2$ consecutive captured coded frames through the video CS measurement model of equation~\eqref{eq:measurementModel}. In both, Table~\ref{tb:psnr_table} and Figure~\ref{fig:allAlgorithmComparison}, results are divided in two parts. The first part lists reconstruction performance of the tested approaches without the MMLE step, while the second compares the performance of the best candidate in the proposed and previous methods, respectively, with a subsequent MMLE step~\cite{Yang2015}. In Table~\ref{tb:psnr_table} the best performing algorithms are highlighted for each part while the bottom row presents average reconstruction time requirements for the recovery of $16$ video frames using $1$ captured coded frame.

\begin{figure*}[!t]
\centering
\includegraphics[scale=1]{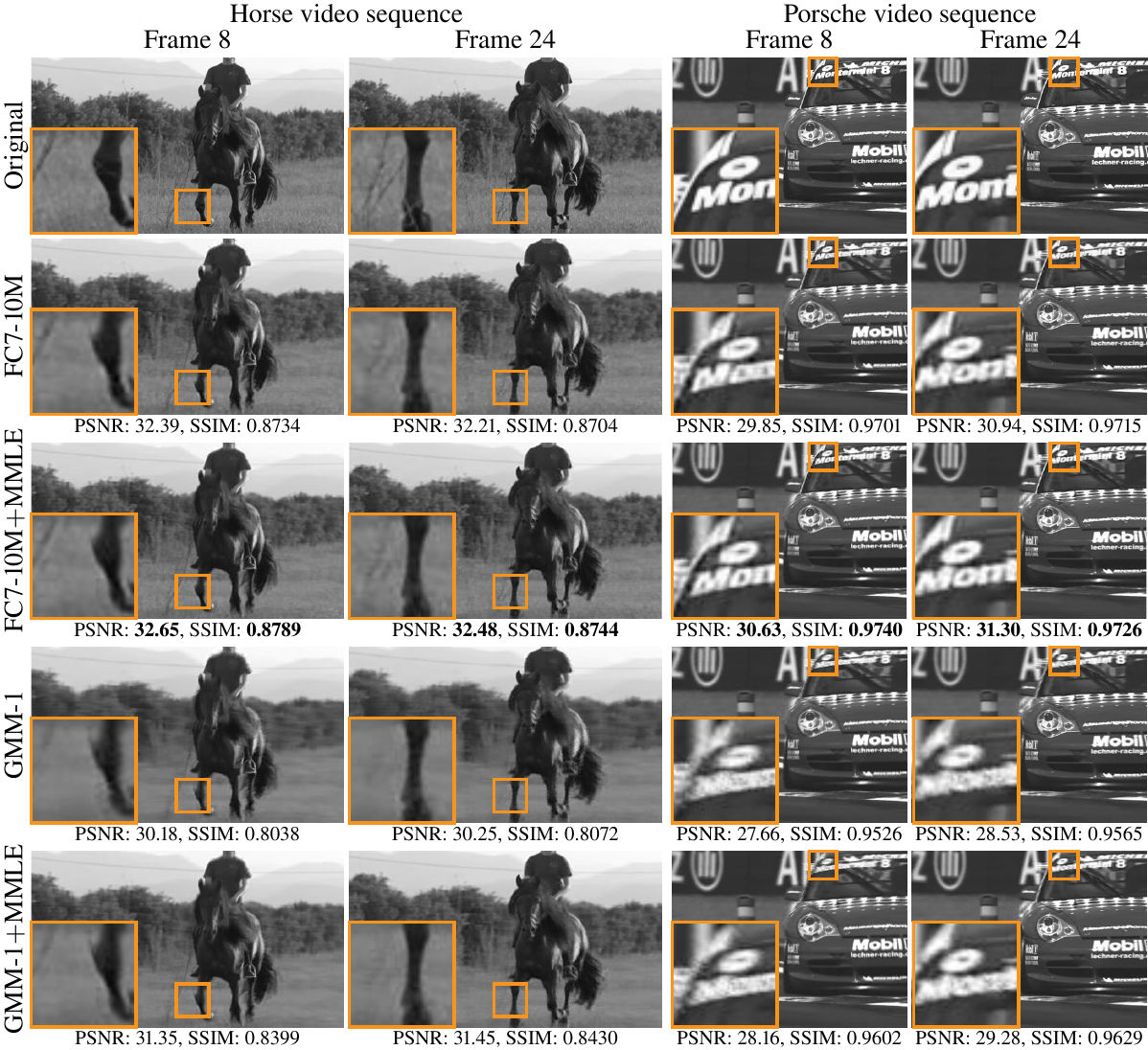}
\caption{Qualitative reconstruction comparison of frames from two video sequences between our methods and GMM-1~\cite{Yang2015}, GMM-1+MMLE~\cite{Yang2015}.}
\label{fig:framesAndInsets}
\end{figure*}

Our FC7-10M and FC7-10M+MMLE yield the highest PSNR and SSIM values for all video sequences. Specifically, the average PSNR improvement of FC7-10M over the GMM-1~\cite{Yang2015} is $2.15$ dB. When these two methods are used to initialize the MMLE~\cite{Yang2015} algorithm, the average PSNR gain of FC7-10M+MMLE over the GMM-1+MMLE~\cite{Yang2015} is $1.67$ dB. Notice also that the FC7-10M achieves $1.01$ dB higher than the combined GMM-1+MMLE. The highest PSNR and SSIM values are reported in the FC7-10M+MMLE method with $33.58$ dB average PSNR over all test sequences. However, the average reconstruction time for the reconstruction of $16$ frames using this method is almost two hours while for the second best, the FC7-10M, is about $12$ seconds, with average PSNR $32.93$ dB. We conclude that, when time is critical, FC7-10M should be the preferred reconstruction method. 

Qualitative results of selected video frames are shown in Figure~\ref{fig:framesAndInsets}. The proposed MLP architectures, including the linear regression model, favorably recover motion while the additional hidden layers emphasize on improving the spatial resolution of the scene (see supplementary material for example reconstructed videos). One can clearly observe the sharper edges and high frequency details produced by the FC7-10M and FC7-10M+MMLE methods compared to previously proposed algorithms. 

\begin{figure*}[!t]
\centering
\includegraphics[scale=1]{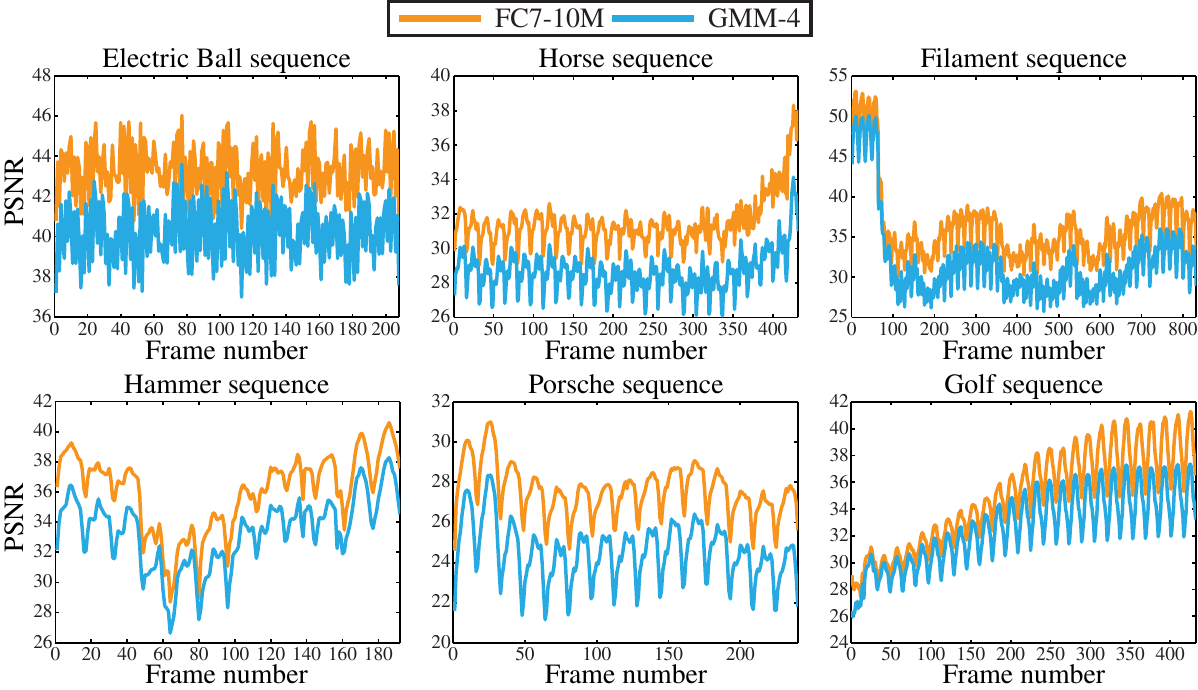}
\caption{PSNR comparison for all the frames of $6$ video sequences between the proposed method FC7-10M and the previous method GGM-4~\cite{Yang2014}.}
\label{fig:fullPSNRComparison}
\end{figure*}

\begin{figure*}[!t]
\centering
\includegraphics[scale=1]{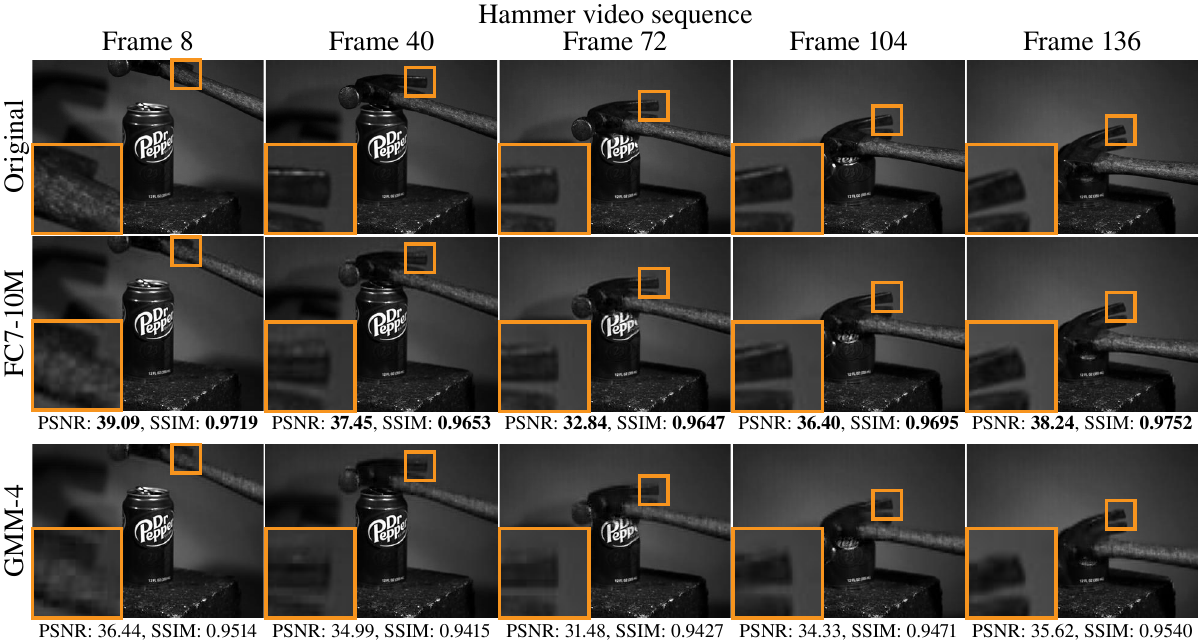}
\caption{Qualitative reconstruction performance of video frames between the proposed method FC7-10M and the previous method GMM-4~\cite{Yang2014}. The corresponding PSNR results for all video frames are shown in Figure~\ref{fig:fullPSNRComparison}.}
\label{fig:framesAndInsetsFullVideo}
\end{figure*}

Due to the extremely long reconstruction times of previous methods, the results presented in Table~\ref{tb:psnr_table} and Figure~\ref{fig:allAlgorithmComparison} refer to only the first $32$ frames of each video sequence, as mentioned above. Figure~\ref{fig:fullPSNRComparison} compares the PSNR for all the frames of $6$ video sequences using our FC7-10M algorithm and the fastest previous method GMM-4~\cite{Yang2014}, while Figure~\ref{fig:framesAndInsetsFullVideo} depicts representative snapshots for some of them. The varying PSNR performance across the frames of a $16$ frame block is consistent for both algorithms and is reminiscent of the reconstruction tendency observed in other video CS papers in the literature~\cite{Koller2015,Llull2013b,Yang2015,Yang2014}.

\subsection{Reconstruction Results with Noise}

Previously, we evaluated the proposed algorithms assuming noiseless measurements. In this subsection, we investigate the performance of the presented deep architectures under the presence of measurement noise. Specifically, the measurement model of equation~\eqref{eq:measurementModel} is now modified to
\begin{equation}
{\bf y} = \Phi {\bf x} + {\bf n},
\end{equation}
where ${\bf n} : M_f \times 1$ is the additive measurement noise vector.

We employ our best architecture utilizing $K = 7$ hidden layers and follow two different training schemes. In the first one, the network is trained on the $10 \times 10^6$ samples, as discussed in subsection~\ref{subsec:comparison} (\ie, the same FC7-10M network as before) while in the second, the network is trained using the same data pairs $\{{\bf y}_i,{\bf x}_i\}$ after adding random Gaussian noise to each vector ${\bf y}_i$. Each vector ${\bf y}_i$ was corrupted with a level of noise such that signal-to-noise ratio (SNR) is uniformly selected in the range between $20-40$ dB giving rise to a set of $10 \times 10^6$ noisy samples for training. We denote the network trained on the noisy dataset as FC7N-10M.

\begin{figure}[!t]
\centering
\includegraphics{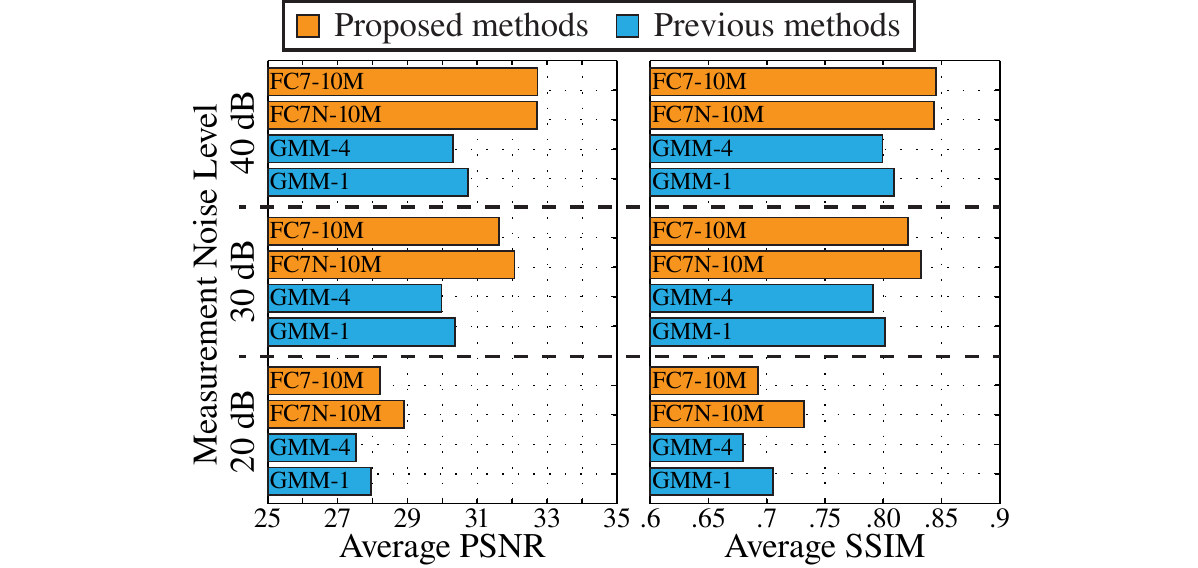}
\caption{Average PSNR and SSIM over all video sequences for several methods under different levels of measurement noise.}
\label{fig:AlgorithmComparisonNoise}
\end{figure}

\begin{figure*}[!t]
\centering
\includegraphics[scale=1]{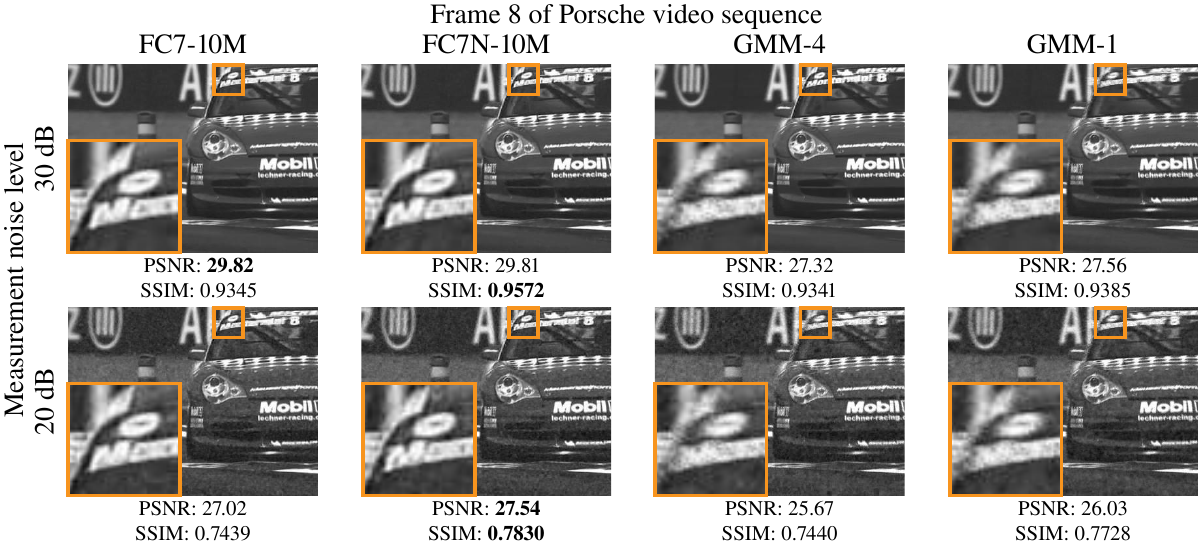}
\caption{Qualitative reconstruction comparison between our methods and GMM-4~\cite{Yang2014}, GMM-1~\cite{Yang2015} under different levels of measurement noise. The original frame and corresponding inset are presented in Figure~\ref{fig:framesAndInsets}.}
\label{fig:AlgorithmComparisonNoiseInsets}
\end{figure*}

We now compare the performance of the two proposed architectures with the previous methods GMM-4 and GMM-1 using measurement noise. We did not include experiments with the MMLE counterparts of the algorithms since, as we observed earlier, the performance improvement is always related to the starting point of the MMLE algorithm. Figure~\ref{fig:AlgorithmComparisonNoise} shows the average performance comparison for the reconstruction of the first $32$ frames of each tested video sequence under different levels of measurement noise while Figure~\ref{fig:AlgorithmComparisonNoiseInsets} depicts example reconstructed frames.

As we can observe, the network trained on noiseless data (FC7-10M) provides good performance for low measurement noise (\eg, $40$ dB) and reaches similar performance to GMM-1 for more severe noise levels (\eg, $20$ dB). The network trained on noisy data (FC7N-10M), proves more robust to noise severity achieving better performance than GMM-1 under all tested noise levels. 

Despite proving more robust to noise, our algorithms in general recover motion favorably but, for high noise levels, there is additive noise throughout the reconstructed scene (observe results for $20$ dB noise level in Figure~\ref{fig:AlgorithmComparisonNoiseInsets}). Such degradation could be combated by cascading our architecture with a denoising deep architecture (\eg,~\cite{Burger2012}) or denoising algorithm to remove the noise artifacts. Ideally, for a specific camera system, data would be collected using this system and trained such that the deep architecture incorporates the noise characteristics of the underlying sensor.

\subsection{Run Time}

Run time comparisons for several methods are illustrated at the bottom row of Table~\ref{tb:psnr_table}. All previous approaches are implemented in MATLAB. Our deep learning methods are implemented in Caffe package~\cite{Jia2014} and all algorithms were executed by the same machine. We observe that the deep learning approaches significantly outperform the previous approaches in order of several magnitudes. Note that a direct comparison between the methods is not trivial due to the different implementations. Nevertheless, previous methods solve an optimization problem during reconstruction while our MLP is a feed-forward network that requires only few matrix-vector multiplications. 

\subsection{Number of Layers and Dataset Size}

From Figure~\ref{fig:allAlgorithmComparison} we observe that as the number of training samples increases the performance consistently improves. However, the improvement achieved by increasing the number of layers (from $4$ to $7$) for architectures trained on small datasets (\eg, 1M) is not significant (performance is almost the same). This is perhaps expected as one may argue that in order to achieve higher performance with extra layers (thus, more parameters to train) more training data would be required. Intuitively, adding hidden layers enables the network to learn more complex functions. Indeed, reconstruction performance in our 10 million dataset is slightly higher in FC7-10M than in FC4-10M. The average PSNR for all test videos is 32.66 dB for FC4-10M and 32.91 dB for FC7-10M. This suggests that 4-hidden layers are sufficient to learn the mappings in our 10M training set. However, we wanted to explore the possible performance benefits of adding extra hidden layers to the network architecture. 

\begin{figure*}[!t]
\centering
\includegraphics[scale=0.90]{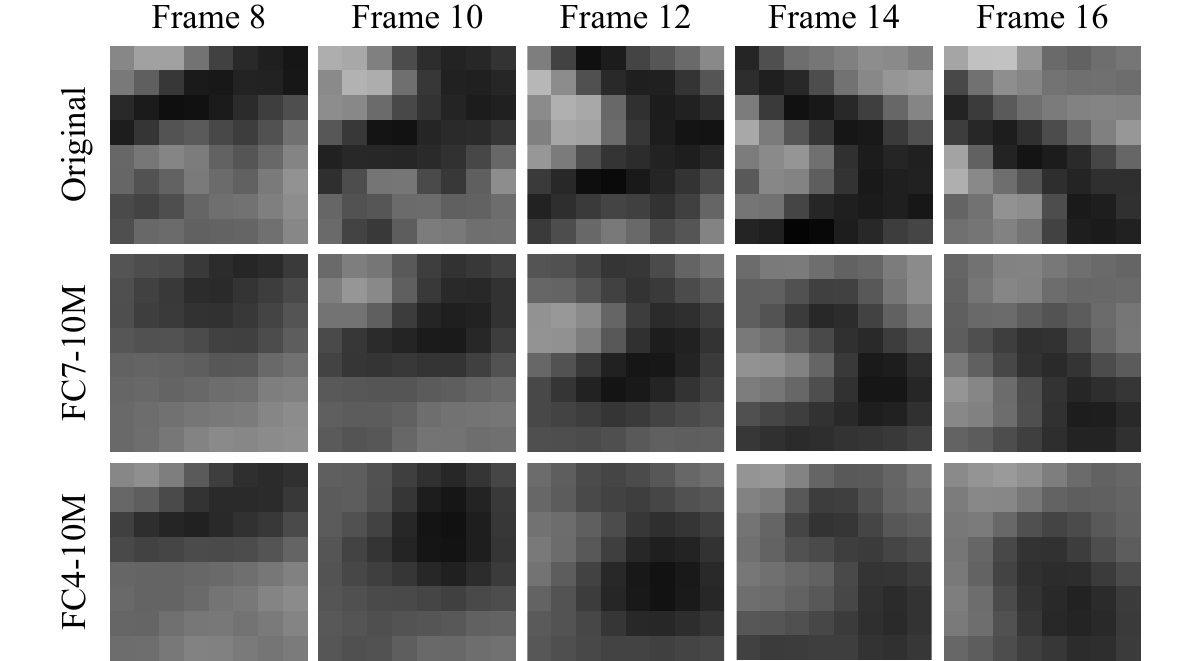}
\caption{Qualitative reconstruction comparison for a video block of the training set. First row shows $5$ patches from the original video block of size $8\times8\times16$; second row shows the reconstruction using the trained network with $7$ hidden layers (FC7-10M); third row shows the reconstruction using the trained network with $4$ hidden layers (FC4-10M). The slight improvement in reconstruction quality using network FC7-10M is apparent while the $\ell_2$ norm reconstruction error is $3.05$ and $4.11$ for FC7-10M and FC4-10M, respectively.}
\label{fig:patch_reconstruction}
\end{figure*}

In order to provide more insights regarding the slight performance improvement of FC7-10M compared to FC4-10M we visualize in Figure~\ref{fig:patch_reconstruction} an example video block from our training set and its respective reconstruction using the two networks. We observe that FC7-10M is able to reconstruct the patches of the video block slightly better than FC4-10M. This suggests that the additional parameters help in fitting the training data more accurately. Furthermore, we observed that reconstruction performance of our validation set was better in FC7-10M than in FC4-10M. Note that a small validation set was kept for tuning the hyper-parameters during training and that we also employed weight regularization ($\ell_2$ norm) to prevent overfitting. Increasing the number of hidden layers further did not help in our experiments as we did not observe any additional performance improvement based on our validation set. Thus, we found that learning to reconstruct training patches accurately was important in our problem.

\section{Conclusions}
 To the best of our knowledge, this work constitutes the first deep learning architecture for temporal video compressive sensing reconstruction. We demonstrated superior performance compared to existing algorithms while reducing reconstruction time to a few seconds. At the same time, we focused on the applicability of our framework on existing compressive camera architectures suggesting that their commercial use could be viable. We believe that this work can be extended in three directions: 1) exploring the performance of variant architectures such as RNNs, 2) investigate the training of deeper architectures and 3) finally, examine the reconstruction performance in real video sequences acquired by a temporal compressive sensing camera.

\balance
{\small
\bibliographystyle{ieee}
\bibliography{videocs}
}

\end{document}